\documentclass[conference]{IEEEtran}
\IEEEoverridecommandlockouts
\usepackage{cite}
\usepackage{amsmath,amssymb,amsfonts}
\usepackage{algorithmic}
\usepackage{graphicx}
\usepackage{textcomp}
\usepackage{xcolor}
\usepackage{booktabs}
\usepackage{arydshln}
\usepackage{booktabs}
\usepackage{tabularx}
\usepackage{url}
\usepackage{colortbl}
\usepackage{soul} 

\def\BibTeX{{\rm B\kern-.05em{\sc i\kern-.025em b}\kern-.08em
    T\kern-.1667em\lower.7ex\hbox{E}\kern-.125emX}}
\begin{document}

\title{Impact of Speech Mode in \\ Automatic Pathological Speech Detection\\
 
}

\author{\IEEEauthorblockN{Shakeel A.~Sheikh, Ina Kodrasi}
\IEEEauthorblockA{\textit{Signal Processing for Communication Group, Idiap Research Institute, Martigny, Switzerland } \\
\{shakeel.sheikh, ina.kodrasi\}@idiap.ch}
}

\maketitle

\begin{abstract}

Automatic pathological speech detection approaches yield promising results in identifying various pathologies. 
These approaches are typically designed and evaluated for phonetically-controlled speech scenarios, where speakers are prompted to articulate identical phonetic content. 
While gathering controlled speech recordings can be laborious, spontaneous speech can be conveniently acquired as potential patients navigate their daily routines. 
Further, spontaneous speech can be valuable in
detecting subtle and abstract cues of pathological speech.
Nonetheless, the efficacy of automatic pathological speech detection for spontaneous speech remains unexplored. 
This paper analyzes the influence of speech mode on pathological speech detection approaches, examining two distinct categories of approaches, i.e., classical machine learning and deep learning.
Results indicate that classical approaches may struggle to capture pathology-discriminant cues in spontaneous speech. 
In contrast, deep learning approaches demonstrate superior performance, managing to extract additional cues that were previously inaccessible in non-spontaneous speech.
\end{abstract}


\section{Introduction}
Pathological speech resulting from neurological disorders poses a significant healthcare challenge, where early diagnosis is crucial for delaying progression and for managing the condition. 
Pathological speech conditions such as dysarthria or apraxia of speech are characterized by alterations in speech production, including imprecise articulation, abnormal pitch, or irregular rhythm, making it challenging for patients to effectively convey their intentions through speech~\cite{damico2010handbook, stewart1995speech, darley1969differential, baghai2012automatic}. 
Automatic speech-based diagnostic approaches emerge as cost-effective methods for early detection \cite{damico2010handbook}, especially compared to traditional diagnostic techniques such as Positron emission tomography and Dopamine transporter scans, which are expensive and inaccessible, creating a barrier to timely diagnosis and treatment~\cite{pet, dat}. 
Such approaches provide convenient monitoring of disease progression, reducing the need for costly in-person therapy sessions. 
As a result, various automatic approaches have been proposed for pathological speech detection, broadly falling into two categories: i) those utilizing classical machine learning techniques with handcrafted acoustic features~\cite{narendra2018dysarthric,Joshy22}, and ii) those based on deep learning architectures operating on time-frequency input representations~\cite{janbakhshi2021supervised,VASQUEZCORREA202056, janbakhshi_stft, bhattacharjee2021source, s2023advancing} or self-supervised embeddings~\cite{w2v2_pd,janbakhshi_ua, s2022machine, javanmardi2023wav2vec, s2023stuttering}.
The first category of approaches often employs support vector machines (SVMs) with Mel-frequency cepstral coefficients (MFCCs)~\cite{orozco2015voiced_svm}, openSMILE features~\cite{janbakhshi_ua}, or sparsity-based features~\cite{kodrasi2020spectro}. 
In contrast, the second category of approaches focuses on exploring time-frequency input representations with different network architectures and training paradigms such as long short-term memory networks~\cite{mallela2020raw}, autoencoders (AEs)~\cite{janbakhshi2022adversarial}, adversarial training~\cite{janbakhshi2021supervised}, or convolutional neural networks (CNNs)~\cite{vasquez2017convolutional, janbakhshi2021supervised}.
More recently, approaches in the second category focus on using self-supervised embeddings such as wav2vec2 for automatic pathological speech detection~\cite{w2v2_pd, janbakhshi_ua}.

While both categories of automatic approaches have demonstrated promising results, they are typically developed and assessed in phonetically-controlled, non-spontaneous, speech settings.
 Phonetically-controlled non-spontaneous speech refers to elicited speech, where speakers are prompted to repeat the same words, phrases, or sentences, meticulously designed by clinicians to elicit discernible cues indicative of pathological speech.
In contrast, spontaneous speech reflects natural conversation and can be valuable in detecting subtle and abstract cues of pathological speech.
Furthermore, spontaneous speech mirrors real-world communication and places a higher cognitive load on real-time processing of various speech tasks such as planning, precise sensorimotor execution, and articulation, making it more susceptible to deficits related to pathology~\cite{damico2010handbook}. 
Yet, to the best of our knowledge, the performance of state-of-the-art automatic pathological speech detection approaches on spontaneous speech remains unexplored. Further investigation is required to assess how these approaches fare in more natural settings, where speech is less structured and influenced by diverse factors such as individual variations in communication styles.

This paper analyzes the effect of \textit{speech mode}, i.e., non-spontaneous and spontaneous speech, on the performance of numerous state-of-the-art automatic pathological speech detection approaches.
The evaluation is conducted on two distinct databases, i.e., the Spanish PC-GITA database containing dysarthria recordings from patients with Parkinson's disease~\cite{orozco2014new} and the French MoSpeeDi database~\cite{mospeedi} containing dysarthria recordings from patients with Parkinson's disease or Amytrophic Lateral Sclerosis.
The considered classical machine learning-based approaches include SVMs operating on hand-crafted acoustic features like openSMILE~\cite{eyben2010opensmile}, MFCCs~\cite{orozco2015voiced_svm}, and sparsity-based features~\cite{kodrasi2020spectro}. 
Additionally, the considered deep learning-based approaches include the CNN-based approach proposed in~\cite{vasquez2017convolutional}, the AE-based approach proposed in~\cite{janbakhshi2021supervised}, and the wav2vec2-based approach in~\cite{janbakhshi_ua}.  

\section{Methods}
This section outlines the state-of-the-art pathological speech detection approaches considered in this paper for analyzing the impact of the speech mode on performance.

 \subsection{Input Representation}
 In the following, we describe the different input representations used in the considered approaches.
 
\paragraph*{OpenSMILE}
OpenSMILE features are commonly used as input to classical machine learning approaches~\cite{janbakhshi_ua, kodrasi2020automatic}.
For each utterance, we extract a $6373$--dimensional feature vector using the openSMILE toolkit~\cite{eyben2010opensmile}. 
Following feature extraction, we employ principal component analysis for dimensionality reduction similar to previous works~\cite{kodrasi2020automatic, janbakhshi_ua}.
We retain only the features that explain 95\% of the variance in the training data. 

\paragraph*{MFCCs}
MFCC features are also commonly used as input to classical machine learning approaches~\cite{Joshy22}.
For each utterance, we compute the mean, variance, kurtosis, and skewness of the first $12$ MFCC coefficients, resulting in a $48$--dimensional feature vector as in~\cite{janbakhshi_ua}. 
The MFCC features are extracted using the openSMILE toolkit~\cite{eyben2010opensmile}. 

\paragraph*{Sparsity-based features}
Sparsity-based features have also been used with classical machine learning approaches~\cite{kodrasi2018statistical, kodrasi2020spectro}.
Following the methodology outlined in~\cite{kodrasi2018statistical, kodrasi2020spectro}, we compute  sparsity-based features for each utterance using the shape parameter of a Chi distribution.
The process involves computing the short-time Fourier transform (STFT) using a Hamming window of length $32$ ms and a hop size of $4$ ms. Next, we obtain the maximum likelihood estimate of the shape parameter for the Chi distribution that best models the spectral magnitude at each frequency bin. This procedure results in a 257-dimensional feature vector for each utterance.

\paragraph*{Mel spectrograms}
 Mel spectrograms are typically used with deep learning-based approaches~\cite{Joshy22}.
For each speech segment, we compute the STFT coefficients using a Hamming window of length $32$~ms and a hop size of $4$~ms.
The computed coefficients are converted into Mel-scale representations with $126$ Mel-bands, following the approach used in~\cite{janbakhshi_ua}.
The logarithm of the Mel spectrogram is then used as the input representation for deep learning-based approaches. 
\paragraph*{Self-supervised embeddings} 
The wav2vec2 framework has revolutionised the extraction of meaningful contextual representations from raw speech~\cite{baevski2020wav2vec}.
These extracted embeddings have shown promising performance in various pathological speech tasks~\cite{vsvec2022evaluation, w2v2_pd, getman2022wav2vec2, s2022end}. 
For our analysis, we use embeddings extracted from the XLSR-53 model~\cite{conneau2020unsupervised}.
It has been shown that the last layers of wav2vec2 models are more adapted towards non paralinguistic and other phonetic content-related tasks, while the first layers are more applicable to paralinguistic and prosody-related tasks~\cite{shah2021all, baevski2020wav2vec, s2022end}.
Hence, in this paper, we consider embeddings extracted from the first transformer layer of the XLSR-53 model.
For each utterance, the final input representation is computed as the mean and standard deviation of the embeddings across time.
This procedure results in a $2048$--dimensional input representation.
It should be noted that the wav2vec2 model is not fine-tuned, but rather utilized as a self-supervised feature extractor for automatic pathological speech detection.
\subsection{Classifiers}
In the following, we describe the different classifiers used in the considered approaches.
\paragraph*{Support vector machines} 
To analyze the performance of classical approaches based on handcrafted acoustic features, we use SVMs with radial basis kernel function as in~\cite{janbakhshi_ua, orozco2015voiced_svm, kodrasi2020spectro}.
Different SVMs are trained for different acoustic features, i.e., for openSMILE, MFCCs, and sparsity-based features.

\paragraph*{Convolutional neural networks}
CNNs have been widely used in various speech applications, including automatic pathological speech detection~\cite{bhattacharjee2021source, javanmardi2024pre, gupta2021residual, akccay2020speech}. 
For our analysis, we train a CNN with Mel spectrogram input representations as in~\cite{janbakhshi_ua}. 
The CNN consists of a normalization layer, followed by two convolutional layers with $64$ channels each and kernel sizes of $2 \times 2$ and $3 \times 3$ respectively. 
Each convolutional layer is followed by batch normalisation, max-pooling with a $2 \times 2$ or $3 \times 3$  kernel, and a ReLU activation function. 
A dropout rate of 30\% is applied after the second convolutional layer. 
Finally, the output is fed to a fully-connected linear layer (input size: $25600$, output size: $2$) for pathological speech detection.

\paragraph*{Autoencoders}
Given the promising performance of the AE-based framework for pathological speech detection in~\cite{janbakhshi2021supervised}, we also consider this approach in our analysis.
Also in this framework  Mel spectrograms are used as the input representation.
The encoder $\theta_e$ is composed of $4$ convolutional layers with a kernel size of $3 \times 3$. 
Each convolutional layer is followed by batch normalization, max-pooling with a $2 \times 2$ kernel, and a ReLU activation function. 
The output of the encoder is fed to a bottleneck  layer of size $128$. 
This bottleneck representation is then decoded by the decoder $\theta_d$ to reconstruct the input representation.
The decoder mirrors the encoder components in reverse order, using interpolation and transposed convolution instead of max-pooling and convolution.
In addition, the bottleneck representation is simultaneously fed to a classifier aiming to learn a pathology-discriminant representation in a multi-task learning fashion. 
The pathology classifier $\theta_{pa}$ is composed of a fully connected linear layer with $128$ input units and $2$ output units. 
The classification loss $\mathcal{L}_{pa}$ and the AE reconstruction loss $\mathcal{L}_{ae}$ are jointly minimized by optimizing the parameters $\theta_e$, $\theta_d$, $\theta_{pa}$ using
\begin{equation}
 \mathcal{L}(\hat \theta_{e}, \hat \theta_{d}, \hat \theta_{pd}) = \underset{\theta_{e}, \theta_{d}, \theta_{pd}}{\mathrm{argmin}} ~ \mathcal{L}(\theta_{e}, \theta_{d}, \theta_{pd}),
\end{equation}
where 
\begin{equation}
    \mathcal{L}( \theta_{e},  \theta_{d},  \theta_{pd}) = 
    (1 - \alpha)\mathcal{L}_{ae}(\theta_{e}, \theta_{d}) + \alpha\mathcal{L}_{pd}(\theta_{e}, \theta_{pd}),
\end{equation}
with $\alpha$ $\in [0, 1]$ being a trade-off parameter between the AE loss and the classification loss. 
In our analysis, we use $\alpha = 0.01$ as in~\cite{janbakhshi_ua}. 
For inference, the decoder is disregarded and the bottleneck representation is directly fed to the pathology classifier. 

\paragraph*{Fully connected linear layer} 
Similar to~\cite{janbakhshi_ua}, we employ a fully connected linear layer for pathological speech classification when using wav2vec2 embeddings as the input representation.
The linear layer consists of $2048$ input units and $2$ output units.

\section{Experimental Settings}
 In this section, the settings used for the experimental analysis are provided.
 \subsection{Databases}
 Analysis are conducted using two different databases, i.e., the Spanish PC-GITA database~\cite{orozco2014new} and the French MoSpeeDi database~\cite{mospeedi}.

\paragraph*{PC-GITA} 
The PC-GITA database contains Spanish recordings from a gender-balanced group of $50$ patients diagnosed with Parkinson’s disease along with a gender-balanced group of $50$ neurotypical speakers.
For the non-spontaneous speech mode, we use recordings of $10$ sentences and a phonetically balanced text. 
Using these recordings, the average length of the total available non-spontaneous speech material across speakers is $55.4$ s.
For the spontaneous speech mode, we use recordings of the speakers talking about what they do in a normal day. 
Using these recordings, the average length of the available spontaneous speech material across speakers is $47.1$ s.
\paragraph*{MoSpeeDi}
From the French MoSpeeDi database, we use recordings from a gender-balanced group of $35$ patients diagnosed with Parkinson’s disease or Amytrophic Lateral Sclerosis along with a gender-balanced group of $35$ neurotypical speakers.
For the non-spontaneous speech mode, we use recordings of $8$ sentences. 
Using these recordings, the average length of the total available non-spontaneous speech material across speakers is $97.7$ s.
For the spontaneous speech mode, we use recordings of the speakers talking about their holidays. 
Using these recordings, the average length of the available spontaneous speech material across speakers is $153.1$ s.

 \subsection{Evaluation and Training} 
 
 For all considered approaches, evaluation is done within a stratified $K$-fold cross validation strategy, with $K=10$ for the PC-GITA database and $K=7$ for the MoSpeeDi database. 
 For each fold of the PC-GITA database, we use $80\%$, $10\%$, and $10\%$ of the data for training, validation, and testing, respectively.
 For each fold of the MoSpeeDi database, we use $72\%$, $14\%$, and $14\%$ of the data for training, validation, and testing, respectively.

 The SVM-based and the wav2vec2-based approaches accept variable length segments of speech as input, hence, full utterances are used as input to these approaches.
 The CNN-based and the AE-based approach accept only fixed-size segments of speech as input.
 For these two approaches, we segment available utterances into $500$ ms segments with a $50\%$ overlap and use these fixed-size segments as input.
 
 The performance is evaluated in terms of speaker-level accuracy.
 For the SVM-based approaches, speaker-level accuracy is computed through majority voting of the decisions for all utterances belonging to each speaker.
 For the remaining approaches, speaker-level accuracy is computed through soft voting of the probability of decisions for all segments/utterances belonging to each speaker.

Implementation is done using PyTorch and Torchaudio.
For the SVM-based approaches, separate SVMs are trained for each of the handcrafted acoustic features, i.e., for openSMILE, MFCCs, and sparsity-based features.
By optimizing the performance on the validation set, the optimal kernel width $\gamma$ and soft margin constant $\mathcal{C}$ is found from the sets $\{10^{-4}, 10^{-3}, 10^{-2}, 10^{-1}]$ and $\{10, 10^2, 10^3, 10^4\}$ respectively. 
The remaining approaches are trained using the Adam optimizer. 
The initial learning rates are $0.0015$ for the AE-based approach, $0.001$ for the CNN-based approach, and $0.001$ for the wav2vec2-based approach.
The learning rates are reduced after every $15$ epochs using the \emph{MultiStepLR}  scheduler with $\gamma$ = 0.9. 
Training is terminated if the validation loss does not decrease for $10$ consecutive epochs, with the last best model saved and used for testing.

\section{Experimental Results}
\begin{table*}[t!] 
\caption{Speaker-level classification accuracy [\%] of the considered automatic pathological speech detection approaches on the PC-GITA and MoSpeeDi database using non-spontaneous and spontaneous speech recordings.}  
\label{tab:pcgita}
   \begin{tabularx}{\linewidth}{Xcccc} 
    \toprule
        Approach & \multicolumn{2}{c}{PC-GITA} & \multicolumn{2}{c}{MoSpeeDi} \\
        & Non-spotaneous & Spontaneous & Non-spotaneous & Spontaneous \\
        \midrule
        SVM with openSMILE input features & 43.0{\color{gray!95}~$\pm$~17.0} & 57.0{\color{gray!95}~$\pm$~17.0}  & 42.9{\color{gray!95}~$\pm$~7.76} & 57.1{\color{gray!95}~$\pm$~19.8} \\
        SVM with MFCC input features  &68.0{\color{gray!95}~$\pm$~09.2} &62.0{\color{gray!95}~$\pm$~13.2} &71.4{\color{gray!95}~$\pm$~10.7} &84.3{\color{gray!95}~$\pm$~15.1}\\
        SVM with sparsity-based input features &73.0{\color{gray!95}~$\pm$~09.5} &56.0{\color{gray!95}~$\pm$~22.2} &82.8{\color{gray!95}~$\pm$~11.1} &78.6{\color{gray!95}~$\pm$~09.0}\\
        \midrule
        CNN with Mel spectrogram input representations &75.0{\color{gray!95}~$\pm$~07.0} &74.0{\color{gray!95}~$\pm$~12.6} &90.1{\color{gray!95}~$\pm$~11.5} &92.9{\color{gray!95}~$\pm$~11.1} \\
        AE with Mel spectrogram input representations &71.0{\color{gray!95}~$\pm$~13.7} &80.0{\color{gray!95}~$\pm$~14.1}&82.9{\color{gray!95}~$\pm$~16.0} &85.7{\color{gray!95}~$\pm$~05.3} \\
        \midrule
        Linear layer with wav2vec2 embedding input representations &77.0{\color{gray!95}~$\pm$~14.9} &75.0{\color{gray!95}~$\pm$~15.8} &87.1{\color{gray!95}~$\pm$~11.12} &87.1{\color{gray!95}~$\pm$~11.1} \\
    
        \bottomrule
  \end{tabularx}
\end{table*}

In this section, we analyse and discuss the impact of speech modes (i.e., non-spontaneous and spontaneous) in SVM-based approaches, CNN- and AE-based approaches, and the wav2vec2-based approach. 
Results obtained for all considered approaches and both databases are presented in Table~\ref{tab:pcgita}.

\subsection*{SVM-based approaches}
As shown in Table~\ref{tab:pcgita}, the different SVM-based approaches do not consistently outperform in one speech mode over the other.
On the one hand, using openSMILE features yields a better performance on spontaneous speech compared to non-spontaneous speech, regardless of the database. 
However, openSMILE features are designed for various tasks like speech emotion recognition or audio detection, which is not ideal for pathological speech detection. Consequently, many openSMILE features capture cues unrelated to pathology. 
This is reflected in the overall low performance achieved by openSMILE features, regardless of the speech mode or database used.
On the other hand, MFCC features demonstrate better performance on spontaneous speech than on non-spontaneous speech in the MoSpeeDi database, while the opposite trend is observed in the PC-GITA database.
In spontaneous speech, the various acoustic, linguistic, spectral, and phonetic features exhibit a lot more non-pathology related fluctuations than in non-spontaneous settings. 
It is expected that MFCC features inadvertently capture phonetic variations in spontaneous recordings, resulting in lower performance for spontaneous speech than for non-spontaneous speech in the PC-GITA database. The absence of this performance trend in the MoSpeeDi database warrants further investigation.
Finally, using sparsity-based features yields a better performance on non-spontaneous speech than on spontaneous speech, regardless of the database.
This result is expected, given that sparsity-based features are derived by fitting a distribution to the speech spectral coefficients. The use of single utterances for this fitting process influences the quality of the fit, leading to distributions that are greatly influenced by the phonetic content of the utterance. The fluctuating phonetic content among spontaneous speech recordings results in unreliable distribution fits that not only exhibit pathological cues, but also phonetic content cues.

In summary, results show that the performance of different SVM-based approaches is differently influenced by the speech mode, depending on the used handcrafted acoustic features.
These findings emphasize the importance of carefully crafting and selecting acoustic features for classical approaches, such that undesirable cues are not captured.

\subsection*{CNN-based and AE-based approaches}

While SVM-based approaches are not be powerful enough to ignore pathology-unrelated fluctuations in spontaneous speech, the CNN-based and AE-based approaches have a better potential. 
Results in Table~\ref{tab:pcgita} show that the CNN-based approach yields a similar performance for both speech modes in both databases whereas the AE-based approach performs better on spontaneous speech recordings than on non-spontaneous ones.
 
These findings indicate that these approaches can not only disregard pathology-unrelated fluctuations in spontaneous speech, but can also extract additional cues that were not identifiable in non-spontaneous speech.
It should be noted that unlike the SVM-based approaches operating on a single utterance, the CNN- and AE-based approaches operate on multiple segments of speech, even for the spontaneous speech recordings. 
The availability of more samples when using the spontaneous speech recordings for the CNN- and AE-based approaches might contribute to the observed improved performance.

\subsection*{wav2vec2-based approach} \enspace
Similar to the CNN- and AE-based approaches, we anticipate the wav2vec2-based approach to be able to disregard pathology-unrelated variations in spontaneous speech. 
As expected, the results in Table~\ref{tab:pcgita} demonstrate that the wav2vec2-based approach performs similarly for both speech modes and databases, with a negligible decrease in performance for spontaneous speech in the PC-GITA database. The XLRS53 variant of the wav2vec2 model is trained on a diverse range of non-spontaneous and spontaneous multilingual databases~\cite{conneau2020unsupervised}, resulting in embeddings that do not exhibit significant fluctuations caused by the speech mode.

\vspace{0.1cm}

\noindent 
Overall, the results presented in this section show that on the one hand, SVM-based approaches struggle to perform well in spontaneous speech recordings depending on the used acoustic features. 
On the other hand, deep learning-based approaches not only have the ability to disregard pathology-unrelated fluctuations in spontaneous speech, but can also extract additional pathology-discriminant cues that may not be present in non-spontaneous speech.
In the future, we will investigate how the performance of the SVM-based and wav2vec2-based approaches changes when the spontaneous speech recording is segmented into shorter fixed-sized segments (as for the CNN- and AE-based approaches), rather than treating it as a single utterance.
Furthermore, we will investigate the extraction of phonetic-dependent acoustic features from spontaneous speech to be used in classical approaches.
 
\section{Conclusion}
Pathological speech detection approaches developed so far have been trained and evaluated only in phonetically-controlled non-spontaneous speech settings. 
However, spontaneous speech can be more conveniently recorded and it can be valuable in detecting subtle and abstract cues of pathological speech.
 In this paper, we have examined the performance of various classical and deep learning-based pathological speech detection approaches using both non-spontaneous and spontaneous speech recordings. Our findings suggest that classical approaches employing handcrafted acoustic features may encounter challenges in capturing pathology-discriminant cues in spontaneous speech. Conversely, deep learning approaches exhibit superior performance, successfully extracting additional cues that were previously inaccessible in non-spontaneous speech.
Future work will target further improving the performance of different approaches on spontaneous speech recordings.

\section{Acknowledgements}
This work was supported by the Swiss National Science Foundation project CRSII5\_202228 on ``Characterisation of motor speech disorders and processes''.

{\footnotesize \bibliographystyle{IEEEtran}
 \bibliography{ref.bib}  }

\end{document}